\documentclass{article}
\usepackage{spconf,amsmath,graphicx}
\usepackage{hyperref}

\title{Representation Mixing for TTS Synthesis}
\name{Kyle Kastner, Jo\~{a}o Felipe Santos, Yoshua Bengio$^{\dagger}$, Aaron Courville${^\star}$\thanks{${\dagger}$ CIFAR Senior Fellow ${^\star}$ CIFAR Fellow}}
\address{MILA\\
	Universit\'{e} de Montr\'{e}al\\
	Montr\'{e}al, Canada}
\begin{document}
%
\maketitle
\begin{abstract}
Recent character and phoneme-based parametric TTS systems using deep learning have shown strong performance in natural speech generation. However, the choice between character or phoneme input can create serious limitations for practical deployment, as direct control of pronunciation is crucial in certain cases. We demonstrate a simple method for combining multiple types of linguistic information in a single encoder, named \emph{representation mixing}, enabling flexible choice between character, phoneme, or mixed representations during inference. Experiments and user studies on a public audiobook corpus show the efficacy of our approach.
\end{abstract}

\section{Introduction}
\label{sec:format}
TTS synthesis \cite{hunt1996unit} focuses on building systems which can generate speech (often in the form of an audio feature sequence $a$), given a set of linguistic sequences, $l$.
These sequences, $l$ and $a$, are of different length and dimensionality thus it is necessary to find the alignment between the linguistic information and the desired audio.
We approach the alignment problem by jointly \emph{learning} to align these two types of information \cite{bahdanau2014neural}, effectively translating the information in the linguistic sequence(s) into audio through learned transformations for effective TTS synthesis \cite{sotelo2017char2wav,wang2017tacotron,shen2017natural,tachibana2018efficiently,ping2018clarinet}. 

\subsection{Data Representation}
 We employ \emph{log mel spectrograms} as the audio feature representation, a well-studied time-frequency representation \cite{smith2011spectral} for audio sequence $a$. Various settings used for this transformation can be seen in Table \ref{table:hyperparameters}. 

Linguistic information, $l$, can be given at the abstract level as graphemes (also known as characters when using English) or at a more detailed level which may include pronunciation information, such as phonemes. Practically, character-level information is widely available in open data sets though this representation allows ambiguity in pronunciation \cite{librivox10}.



\subsection{Motivating Representation Mixing}
In some cases, it may be impossible to fully realize the desired audio without being given direct knowledge of pronunciation. Take as a particular example the sentence \emph{"All travelers were exhausted by the wind down the river"}. The word \emph{"wind"} in this sentence can be either noun form such as \emph{"The wind blew swiftly on the plains"}, or verb form for traveling such as \emph{"... and as we wind on down the road"}, both of which have different pronunciation.

Without external knowledge (such as additional context, or an accompanying video) of which pronunciation to use for the word, a TTS system which operates on character input will always have ambiguity in this situation. Alternate approaches such as grapheme to phoneme methods \cite{rao2015grapheme} will also be unable to resolve this problem. Such cases are well-known in both TTS and linguistics \cite{black1998issues,eddington2015meaning}, motivating our desire to flexibly combine grapheme and phoneme inputs in a single encoder. Our method allows per example control of pronunciation during inference \emph{without requiring} this information for all desired inputs, and similar methods have also been important for other systems \cite{ping2018deep}.
\begin{figure}[htb]
\begin{minipage}[b]{\linewidth}
  \centering
  \centerline{\includegraphics[width=8cm,height=4.1cm]{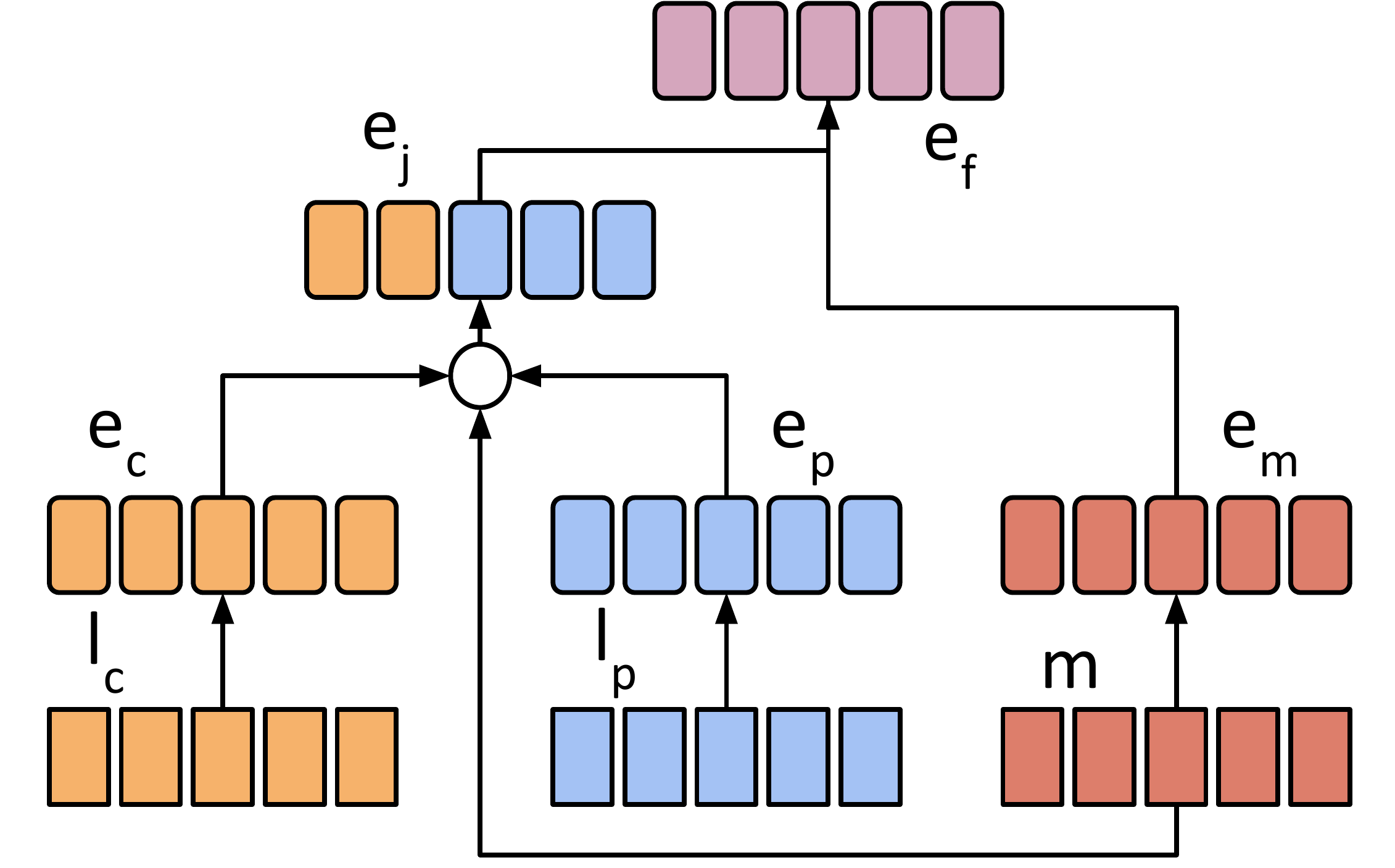}}
\end{minipage}
\caption{\small{Visualization of embedding computation}}
\label{fig:embedding}
\end{figure}
\vspace{-.5cm}

\section{Representation Mixing Description}
The input to the system consists of one data sequence, $l_j$, and one mask sequence, $m$. The data sequence, $l_j$, consists of a mixture between a character sequence, $l_c$, and a phonetic sequence, $l_p$. The mask sequence describes which respective sequence the symbol came from with an integer ($0$ or $1$). 
Each time a training sequence is encountered, it is randomly mixed at the word level and we assign all spaces and punctuation as characters.

For example, a pair "$the$ $cat$", "$@d@ah$ $@k@ah@t$" (where @ is not a used symbol and serves only as an identifier to mark phoneme boundaries) can be resampled as "$the$ $@k@ah@t$", with a corresponding mask of $[0, 0, 0, 0, 1, 1, 1]$. 
On the next occurrence, it may be resampled as "$@d@ah$ $cat$", with mask $[1, 1, 0, 0, 0, 0]$.  
This resampling procedure can be seen as data augmentation, as well as training the model to smoothly process both character and phoneme information without over-reliance on either representation. We demonstrate the importance of each aspect in Section \ref{sec:experiments}.

\subsection{Combining Embeddings}
The full mixed sequence, $l_j$, separately passes through two embedding matrices, $e_c$ and $e_p$, and is then combined using the mask sequence to form a joint mixed embedding, $e_j$. For convenience, $e_c$ and $e_p$, are set to a vocabulary size that is the \emph{max} of the character and phoneme vocabulary sizes. This mixed embedding is further combined with an embedding of the mask sequence itself, $e_m$, for a final combined embedding, $e_f$. This embedding, $e_f$, is treated as the standard input for the rest of the network. A diagram describing this process is shown in Figure \ref{fig:embedding}. 
\begin{align}
    e_j =&\: (1 - m) * e_c + m * e_p \\
    e_f =&\: e_m + e_j
\end{align}

\subsection{Stacked Multi-scale Residual Convolution}
\begin{figure}[htb]
\begin{minipage}[b]{\linewidth}
  \centering
  \centerline{\includegraphics[width=8.0cm,height=5.5cm]{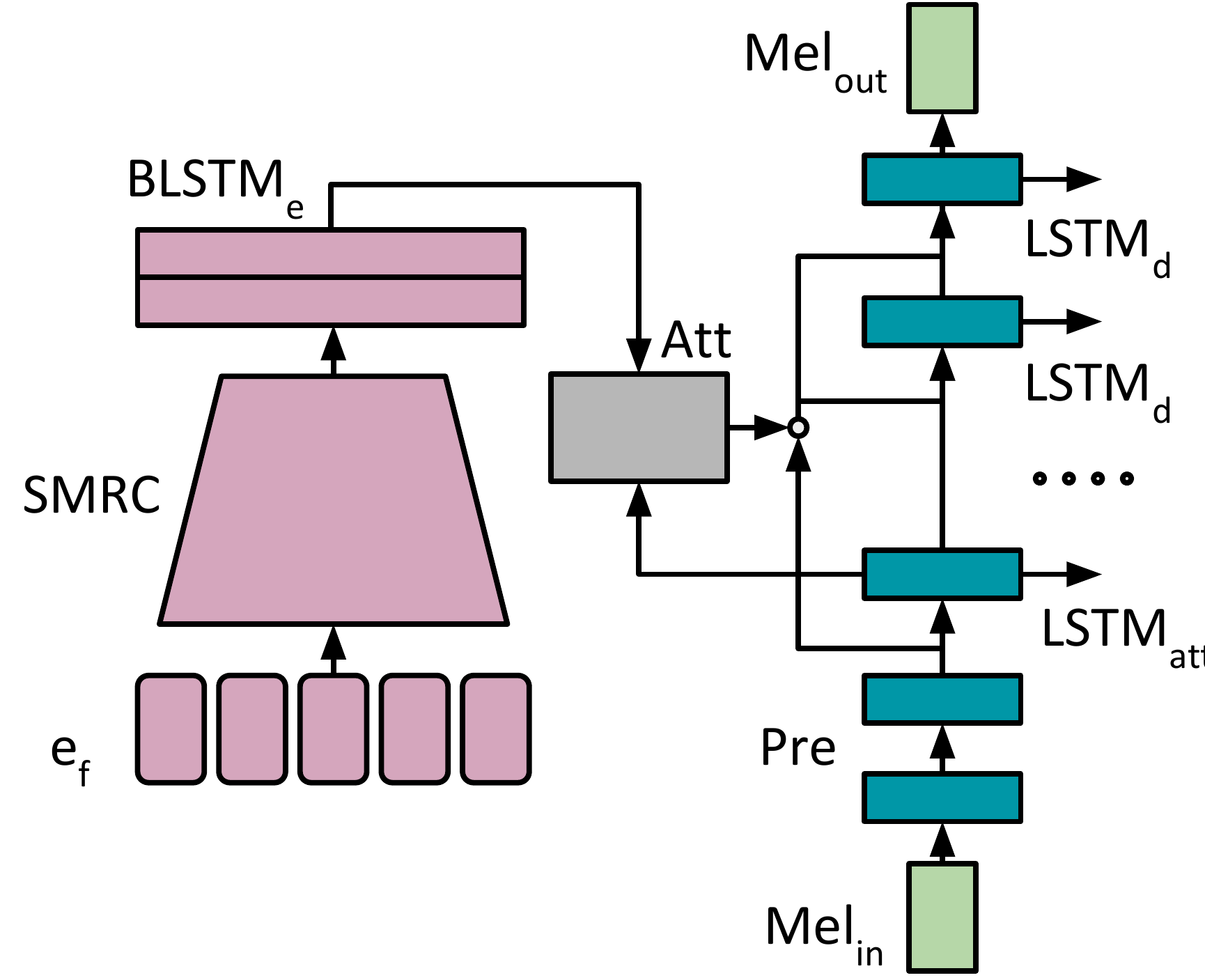}}
\end{minipage}
\caption{\small{Encoding, attention, and one step of mel decoder}}
\label{fig:network}
\end{figure}

In the next sections, we describe the network architecture for transforming the mixed representation into a spectrogram and then an audio waveform.  A full system diagram of our neural network architecture can be seen in Figure \ref{fig:network}.

The final embedding, $e_f$, from the previous section is used as input to a stacked multi-scale residual convolutional subnetwork (SMRC). The SMRC consists of several layers of multi-scale convolutions, where each multi-scale layer is in turn a concatenation of $1\times1$, $3\times3$, and $5\times5$ layers concatenated across the channel dimension. The layers are connected using residual bypass connections \cite{heres2016} and batch normalization \cite{Ioffe+Szegedy-2015} is used throughout. After the convolutional stage, the resulting activations are input to a bidirectional LSTM layer \cite{schuster1997bidirectional,Hochreiter+Schmidhuber-1997}. This ends the \emph{encoding} part of the network.
\subsection{The Importance of Noisy Teacher Forcing}
All audio information in the network passes through a multilayer pre-net with dropout \cite{srivastava2014dropout,wang2017tacotron,shen2017natural}, in both training and evaluation. We found that using pre-net corrupted audio in every layer (including the attention layer) greatly improves the robustness of the model at evaluation, while corruption of linguistic information made the generated audio sound less natural.
%

\subsection{Attention-based RNN Decoder}
The encoding activations are attended using a Gaussian mixture (GM) attention \cite{Graves2013} driven by an LSTM network (conditioned on both the text and pre-net activations), with a softplus activation for the step of the Gaussian mean as opposed to the more typical exponential activation. We find that a softplus step substantially reduces instability during training. This is likely due to the relative length differences between linguistic input sequences and audio output.

Subsequent LSTM decode layers are conditioned on pre-net activations, attention activations, and the hidden state of the layer before, forming a series of skip connections \cite{Graves2013,zhang2016architectural}. LSTM decode layers utilize cell dropout regularization \cite{semeniuta2016recurrent}. The final hidden state is projected to match the dimensionality of the audio frames and a mean squared error loss is calculated between the predicted and true next frames. 

\subsection{Truncated Backpropagation Through Time (TBPTT)}
\begin{figure}[htb]
\begin{minipage}[b]{\linewidth}
  \centering
  \centerline{\includegraphics[width=7.0cm]{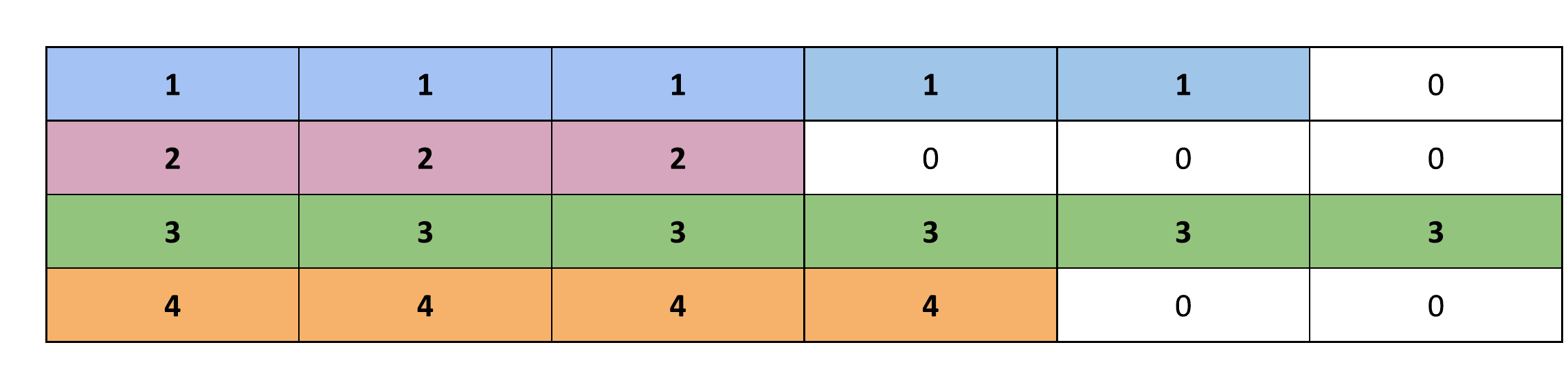}}
  \vspace{-.3cm}
  \centerline{\small{Single minibatch}}
\end{minipage}
\hfill
\begin{minipage}[b]{\linewidth}
  \centering
  \centerline{\includegraphics[width=8.0cm]{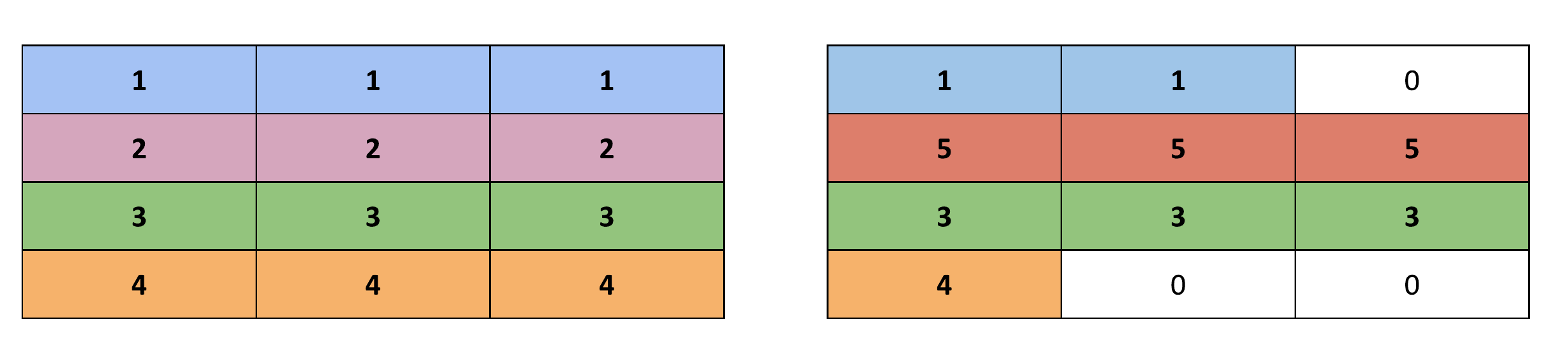}}
  \vspace{-.3cm}
  \centerline{\small{Truncated minibatches with resets}}
\end{minipage}
\vspace{-.6cm}
\caption{\small{Comparison of standard minibatching versus truncation.}}
\label{fig:minibatch}
\end{figure}

The network uses truncated backpropagation through time (TBPTT) \cite{Hochreiter+Schmidhuber-1997} in the decoder, only processing a subsection of the relevant audio sequence while reusing the linguistic sequence until the end of the associated audio sequence. TBPTT training allows for a particular iterator packing scheme not available with full sequence training, which continuously packs subsequences resetting only the particular elements of the sequence minibatches (and the associated RNN state) when reaching the end of an audio sequence. A simplified diagram of this approach can be seen in Figure \ref{fig:minibatch}.


\subsection{Converting Features Into Waveforms}
After predicting a log mel spectrogram sequence from linguistic input, further transformation must be applied to get an audible waveform. A variety of methods have been proposed for this inversion such as the Griffin-Lim transform \cite{griffin1984signal}, and modified variants \cite{slaney1994auditory}. 
Neural decoders such as the conditional WaveNet decoder \cite{vandenoordwavenet2016,shen2017natural} or conditional SampleRNN \cite{mehri2016samplernn} can also act effectively as an inversion procedure to transform log mel spectrograms into audible waveforms.

Optimization methods also work for inversion and we utilize an L-BFGS based routine. 
This process optimizes randomly initialized parameters (representing the desired waveform) passed through a log mel spectrogram transform, to make the transform of these parameters match a given log mel spectrogram \cite{Thome-personal-communication}.

This results in a set of parameters (which we now treat as a fixed waveform) with a lossy transform that closely matches the given log mel spectrogram. The overall procedure closely resembles effective techniques in other generative modeling settings \cite{gatys2016image}.
\begin{table}[htb]
    \begin{center}
    \begin{tabular}{ | l | l | l | l |}
    \hline
    \emph{Inversion Method}  & \emph{Seconds per Sample} & \emph{STRTF}  \\ \hline
    100 L-BFGS        & 1.49E-4           & 3.285 \\ \hline
    1000 L-BFGS       & 1.32E-3           & 29.08 \\ \hline
    Modified Griffin-Lim & 2.81E-4           & 6.206 \\ \hline
    100 L + GL        & 4.11E-4           & 9.063 \\ \hline
    1000 L + GL       & 1.6E-3           & 35.28 \\ \hline
    WaveNet           & 7.9E-3           & 174.7 \\ \hline
    100 L + GL + WN   & 8.3E-3           & 183.7 \\ \hline
    1000 L + GL + WN  & 9.5E-3           & 210.0 \\ \hline
    \end{tabular}
    \caption{\small{Comparison of various log mel spectrogram to waveform conversion techniques. STRTF stands for "slower than real time factor" (calculated assuming output sample rate of $22.05$ kHz), where $1.0$ would be real-time generation for a single example.}}
    \label{table:compute}
    \end{center}
\end{table}
\vspace{-1cm}
\subsection{Inversion Pipeline}
This work uses a combined pipeline of L-BFGS based inversion, followed by modified Griffin-Lim for waveform estimation \cite{slaney1994auditory}. The resulting waveform is of moderate quality, and usable for certain applications including speech recognition. Using either L-BFGS or Griffin-Lim in isolation did not yield usable audio in our experiments. We also found that this ordering (L-BFGS then Griffin-Lim) in our two-stage pipeline works much better than the reverse setting.

To achieve the highest quality output, we then use the moderate quality waveform output from the two stage pipeline, converted back to log mel spectrograms, for conditional WaveNet synthesis. Though other work \cite{shen2017natural} clearly shows that log mel spectrogram conditioned WaveNet can be used directly, we find the two stage pipeline allows for quicker quality checking during development, as well as slightly higher quality in the resulting audio after WaveNet sampling. We suspect this is because the pre-trained WaveNet \cite{yamamoto18wavenet} that we use is trained on ground-truth log mel spectrograms rather than predicted outputs. The synthesis speed of each setting is shown in Table \ref{table:compute}. 
\section{Related Work}
Representation mixing is closely related to the "mixed-character-and-phoneme" setting described in Deep Voice 3 \cite{ping2018deep}, with the primary difference being our addition of the mask embedding $e_m$. We found utilizing a mask embedding alongside the character and phoneme embeddings further improved quality, and was an important piece in the text portion of the network. The focused user study in this paper also highlights the advantages of this kind of mixing independent of high-level architecture choices, since our larger system is markedly different from that used in Deep Voice 3.

The SMRC subnetwork is closely related to the CBHG encoder \cite{wang2017tacotron}, differing in the number and size of convolutional scales used, no max-pooling, use of residual connections, and multiple stacks of multi-scale layers.
We use a Gaussian mixture attention, first described by Graves \cite{Graves2013} and used in several speech related papers \cite{sotelo2017char2wav,skerry2018towards,taigman2017loop}, though we utilize \emph{softplus} activation for the mean-step parameter finding that it improved stability in early training.

Our decoder LSTMs utilize cell dropout \cite{semeniuta2016recurrent} as seen in prior work \cite{ha2016hypernetworks,ha2017neural}. Unlike Tacotron \cite{wang2017tacotron}, we do not use multi-step prediction, convolutional layers in the mel-decoding stage, or Griffin-Lim inside the learning pathway. Audio processing closely resembles Tacotron 2 \cite{shen2017natural} overall, though we precede conditional WaveNet with a computationally efficient two stage L-BFGS and Griffin-Lim inference pipeline for quality control, and as an alternative to neural inversion.

\section{Experiments}
\label{sec:experiments}
The model is trained on LJSpeech \cite{ljspeech17}, a curated subset of the LibriVox corpus \cite{librivox10}. LJSpeech consists of $13,100$ audio files (comprising a total time of approximately $24$ hours) of read English speech, spoken by Linda Johnson. The content of these recordings is drawn from scientific, instructional, and political texts published between $1893$ and $1964$. The recordings are stored as $22.05$ kHz, $16$ bit WAV files, though these are conversions from the original $128$ kbps MP3 format.

Character level information is extracted from audio transcriptions after a normalization and cleaning pipeline. This step includes converting all text to lowercase along with expanding acronyms and numerical values to their spelled-out equivalents. Phonemes are then extracted from the paired audio and character samples, using a forced alignment tool such as Gentle \cite{gentle17}.
This results in character and phoneme information aligned along word level boundaries, so that randomly \emph{mixed} linguistic sequences can be sampled repeatedly throughout training. When using representation mixing for training, we choose between characters and phonemes with probability $.5$ for each word in all experiments. 

\begin{table}[h]
    \begin{tabular}{ | p{2.5cm} | p{5.5cm} |}
    \hline
    Audio Processing & \footnotesize{$22.05$ kHz $16$ bit input, scale between $(-1, 1)$, log mel spectrogram $80$ mel bands, window size $512$, step $128$, lower edge $125$ Hz, upper edge $7.8$ kHz, mean and std dev normalized per feature dimension after log mel calculation.}\\ \hline
    Embeddings & \footnotesize{vocabulary size (v) $49$, $49$, $2$, embedding dim $15$, truncated normal $\frac{1}{\sqrt{v}}$} \\ \hline
    SMRC & \footnotesize{$3$ stacks, stack scales $1\times1$, $3\times3$, $5\times5$, $128$ channels each, batch norm, residual connections, \emph{ReLU} activations, orthonormal init} \\ \hline
    Enc Bidir LSTM & \footnotesize{hidden dim $1024$ $(128\times4\times 2)$, truncated normal init scale $0.075$} \\ \hline
    Pre-net & \footnotesize{$2$ layers, hidden dim $128$, dropout keep prob $0.5$, linear activations, orthonormal init} \\ \hline
    Attention LSTM & \footnotesize{num mixes $10$, \emph{softplus} step activation, hidden dim $2048$ $(512\times4)$, trunc norm init scale $0.075$} \\ \hline
    Decoder LSTMs & \footnotesize{$2$ layers, cell dropout keep prob $0.925$, truncated normal init scale $0.075$} \\ \hline
    Optimizer & \footnotesize{mse, no output masking, Adam optimizer, learning rate $1E-4$, global norm clip scale $10$} \\ \hline
    Other & \footnotesize{TBPTT length $256$, batch size $64$, training steps $500$k, $1$ TitanX Pascal GPU, training time $7$ days} \\ \hline
    \end{tabular}
    \caption{\small{Architecture hyperparameter settings}}
    \label{table:hyperparameters}
\end{table}
\vspace{-.5cm}

\subsection{Log Mel Spectrogram Inversion Experiments}
We use a high quality implementation of WaveNet, including a pre-trained model directly from Yamamoto et. al. \cite{yamamoto18wavenet}. This model was also trained on LJSpeech, allowing us to directly use it as a neural inverse to the log mel spectrograms predicted by the attention-based RNN model, or as an inverse to log mel spectrograms extracted from the network predictions after further processing.
Ultimately combining the two stage L-BFGS and modified Griffin-Lim pipeline with a final conditional WaveNet sampling pass demonstrated the best quality, and is what we used for the user study shown in Table \ref{table:study}. 

\subsection{Preference Testing}

Given the overall system architecture, we pose three primary questions (referenced in column Q in Table \ref{table:study}) as a user study:

\vspace{.1cm}

1) Does representation mixing (RM) improve character-based inference over a baseline which was trained only on characters?

2) Does RM improve phoneme with character backoff for unknown word (PWCB) inference over a baseline trained on fixed PWCB? 

3) Does PWCB inference improve over character-based inference in a model trained with representation mixing? 

\vspace{.1cm}

To answer these questions, we pose each as a preference test by presenting study participants with $20$ paired choices. Each user was instructed to choose the sample they preferred \footnote{{\fontsize{6}{6} \url{https://s3.amazonaws.com/representation-mixing-site/index.html}}}
. The $20$ tests were chosen randomly in a pool of $123$ possible tests covering all three question types, from $41$ possible sentences. Presentation order was randomized for each question and every user. The overall study consisted of $22$ users and $429$ responses across all categories (some users didn't select any preference for some questions).
\begin{table}[htb]
    \begin{center}
    \begin{tabular}{ | l | l | l | l | l | l |}
    \hline
    \emph{Q} & \textbf{\emph{Model A}} &  \emph{Model B}  & \emph{C. A} & \emph{Total} & \emph{\% A}\\ \hline
    1 & {RM (char)} & Char   & 101  & 137 & 73.7\% \\ \hline
    2 & {RM (PWCB)} & PWCB   & 106 & 145 & 73.1\%  \\ \hline
    3 & {RM (PWCB)} & RM (char) & 86 & 147 & 58.5\% \\ \hline
    \end{tabular}
    \caption{\small{A / B user preference study results. RM uses approach in parenthesis at inference time. Columns "\emph{C. A}" and "\emph{\% A}"  indicate the number and percentage of users which preferred \emph{Model A}.}}
    \label{table:study}
    \end{center}
\end{table}
\vspace{-.75cm}

We see that users clearly prefer models trained with representation mixing, even when using identical information for inference. This highlights the data augmentation aspects of representation mixing, as regardless of information type representation mixing (RM) gives clearly preferable results compared to static representations (Char, PWCB). The preference of representation mixing over static PWCB also means that introducing phoneme and character information without mixing is less beneficial than full representation mixing.

It is also clear that for representation mixing models, pronunciation information (PWCB) at inference gives preferable samples compared to character information. This is not surprising, but further reinforces the importance of using pronunciation information where possible. Representation mixing enables choice in input format, allowing the possibility to use many different inference representations for a given sentence with a single trained model. 

\section{Conclusion}
\label{sec:print}
This paper shows the benefit of \emph{representation mixing}, a simple method for combining multiple types of linguistic information for TTS synthesis. Representation mixing enables inference conditioning to be controlled independently of training representation, and also results in improved quality over strong character and phoneme baselines trained on a publically available audiobook corpus.


\bibliographystyle{IEEEbib}
{\small \bibliography{joined}}

\begin{thebibliography}{10}

\bibitem{hunt1996unit}
A.~J. Hunt and A.~W. Black,
\newblock ``Unit selection in a concatenative speech synthesis system using a
  large speech database,''
\newblock in {\em Acoustics, Speech, and Signal Processing, 1996. ICASSP-96.
  Conference Proceedings., 1996 IEEE International Conference on}. IEEE, 1996,
  vol.~1, pp. 373--376.

\bibitem{bahdanau2014neural}
D.~Bahdanau, K.~Cho, and Y.~Bengio,
\newblock ``Neural machine translation by jointly learning to align and
  translate,''
\newblock in {\em International Conference on Learning Representations (ICLR
  2015)}, 2015.

\bibitem{sotelo2017char2wav}
J.~Sotelo, S.~Mehri, K.~Kumar, J.~F. Santos, K.~Kastner, A.~Courville, and
  Y.~Bengio,
\newblock ``Char2wav: End-to-end speech synthesis,''
\newblock 2017.

\bibitem{wang2017tacotron}
Y.~Wang, R.~Skerry-Ryan, D.~Stanton, Y.~Wu, R.~J. Weiss, N.~Jaitly, Z.~Yang,
  Y.~Xiao, Z.~Chen, S.~Bengio, et~al.,
\newblock ``Tacotron: A fully end-to-end text-to-speech synthesis model,''
\newblock {\em arXiv preprint}, 2017.

\bibitem{shen2017natural}
J.~Shen, R.~Pang, R.~J. Weiss, M.~Schuster, N.~Jaitly, Z.~Yang, Z.~Chen,
  Y.~Zhang, Y.~Wang, R.~Skerry-Ryan, et~al.,
\newblock ``Natural tts synthesis by conditioning wavenet on mel spectrogram
  predictions,''
\newblock {\em arXiv preprint arXiv:1712.05884}, 2017.

\bibitem{tachibana2018efficiently}
H.~Tachibana, K.~Uenoyama, and S.~Aihara,
\newblock ``Efficiently trainable text-to-speech system based on deep
  convolutional networks with guided attention,''
\newblock in {\em ICASSP}, 2018.

\bibitem{ping2018clarinet}
W.~{Ping}, K.~{Peng}, and J.~{Chen},
\newblock ``{ClariNet: Parallel Wave Generation in End-to-End
  Text-to-Speech},''
\newblock {\em ArXiv e-prints}, July 2018.

\bibitem{smith2011spectral}
J.~O. Smith,
\newblock {\em Spectral Audio Signal Processing},
\newblock online book, 2011 edition.

\bibitem{librivox10}
``Librivox,'' http://librivox.org/.

\bibitem{rao2015grapheme}
K.~Rao, F.~Peng, H.~Sak, and F.~Beaufays,
\newblock ``Grapheme-to-phoneme conversion using long short-term memory
  recurrent neural networks,''
\newblock in {\em ICASSP}, 2015.

\bibitem{black1998issues}
A.~W. Black, K.~Lenzo, and V.~Pagel,
\newblock ``Issues in building general letter to sound rules,''
\newblock 1998.

\bibitem{eddington2015meaning}
C.~M. Eddington and N.~Tokowicz,
\newblock ``How meaning similarity influences ambiguous word processing: The
  current state of the literature,''
\newblock {\em Psychonomic bulletin \& review}, vol. 22, no. 1, pp. 13--37,
  2015.

\bibitem{ping2018deep}
W.~Ping, K.~Peng, A.~Gibiansky, S.~O. Arik, A.~Kannan, S.~Narang, J.~Raiman,
  and J.~Miller,
\newblock ``Deep voice 3: Scaling text-to-speech with convolutional sequence
  learning,''
\newblock in {\em International Conference on Learning Representations (ICLR
  2018)}, 2018.

\bibitem{heres2016}
K.~He, X.~Zhang, S.~Ren, and J.~Sun,
\newblock ``Deep residual learning for image recognition,''
\newblock in {\em Proceedings of the IEEE conference on Computer Vision and
  Pattern Recognition}, 2016.

\bibitem{Ioffe+Szegedy-2015}
S.~Ioffe and C.~Szegedy,
\newblock ``Batch normalization: Accelerating deep network training by reducing
  internal covariate shift,''
\newblock 2015.

\bibitem{schuster1997bidirectional}
M.~Schuster and K.~K. Paliwal,
\newblock ``Bidirectional recurrent neural networks,''
\newblock {\em IEEE Transactions on Signal Processing}, vol. 45, no. 11, pp.
  2673--2681, 1997.

\bibitem{Hochreiter+Schmidhuber-1997}
S.~Hochreiter and J.~Schmidhuber,
\newblock ``Long short-term memory,''
\newblock {\em Neural Computation}, vol. 9, no. 8, pp. 1735--1780, 1997.

\bibitem{srivastava2014dropout}
N.~Srivastava, G.~Hinton, A.~Krizhevsky, I.~Sutskever, and R.~Salakhutdinov,
\newblock ``Dropout: A simple way to prevent neural networks from
  overfitting,''
\newblock {\em The Journal of Machine Learning Research}, vol. 15, no. 1, pp.
  1929--1958, 2014.

\bibitem{Graves2013}
A.~{Graves},
\newblock ``Generating sequences with recurrent neural networks,''
\newblock {\em ar{X}iv:{\tt 1308.0850 [cs.NE]}}, Aug. 2013.

\bibitem{zhang2016architectural}
S.~Zhang, Y.~Wu, T.~Che, Z.~Lin, R.~Memisevic, R.~R. Salakhutdinov, and
  Y.~Bengio,
\newblock ``Architectural complexity measures of recurrent neural networks,''
\newblock in {\em Advances in Neural Information Processing Systems}, 2016, pp.
  1822--1830.

\bibitem{semeniuta2016recurrent}
S.~Semeniuta, A.~Severyn, and E.~Barth,
\newblock ``Recurrent dropout without memory loss,''
\newblock in {\em Proceedings of COLING 2016, the 26th International Conference
  on Computational Linguistics: Technical Papers}, 2016, pp. 1757--1766.

\bibitem{griffin1984signal}
D.~Griffin and J.~Lim,
\newblock ``Signal estimation from modified short-time fourier transform,''
\newblock {\em IEEE Transactions on Acoustics, Speech, and Signal Processing},
  vol. 32, no. 2, pp. 236--243, 1984.

\bibitem{slaney1994auditory}
M.~Slaney, D.~Naar, and R.~Lyon,
\newblock ``Auditory model inversion for sound separation,''
\newblock in {\em Acoustics, Speech, and Signal Processing, 1994. ICASSP-94.,
  1994 IEEE International Conference on}. IEEE, 1994, vol.~2, pp. II--77.

\bibitem{vandenoordwavenet2016}
A.~{van den Oord}, S.~{Dieleman}, H.~{Zen}, K.~{Simonyan}, O.~{Vinyals},
  A.~{Graves}, N.~{Kalchbrenner}, A.~{Senior}, and K.~{Kavukcuoglu},
\newblock ``{WaveNet: A Generative Model for Raw Audio},''
\newblock {\em ArXiv e-prints}, Sept. 2016.

\bibitem{mehri2016samplernn}
S.~Mehri, K.~Kumar, I.~Gulrajani, R.~Kumar, S.~Jain, J.~Sotelo, A.~Courville,
  and Y.~Bengio,
\newblock ``Samplernn: An unconditional end-to-end neural audio generation
  model,''
\newblock {\em arXiv preprint arXiv:1612.07837}, 2016.

\bibitem{Thome-personal-communication}
C.~Thom\'{e},
\newblock ``Personal communication,''
\newblock 2018.

\bibitem{gatys2016image}
L.~A. Gatys, A.~S. Ecker, and M.~Bethge,
\newblock ``Image style transfer using convolutional neural networks,''
\newblock in {\em CVPR}, 2016, pp. 2414--2423.

\bibitem{yamamoto18wavenet}
R.~Yamamoto, M.~Andrews, M.~Petrochuk, W.~Hy, cbrom, O.~Vishnepolski,
  M.~Cooper, K.~Chen, and A.~Pielikis,
\newblock ``r9y9/wavenet\_vocoder: v0.1.1 release,'' Oct. 2018,
\newblock https://github.com/r9y9/wavenet\_vocoder.

\bibitem{skerry2018towards}
R.~Skerry-Ryan, E.~Battenberg, Y.~Xiao, Y.~Wang, D.~Stanton, J.~Shor, R.~J.
  Weiss, R.~Clark, and R.~A. Saurous,
\newblock ``Towards end-to-end prosody transfer for expressive speech synthesis
  with tacotron,''
\newblock {\em arXiv preprint arXiv:1803.09047}, 2018.

\bibitem{taigman2017loop}
Y.~Taigman, L.~Wolf, A.~Polyak, and E.~Nachmani,
\newblock ``Voice synthesis for in-the-wild speakers via a phonological loop,''
\newblock {\em arXiv preprint arXiv:1707.06588}, 2017.

\bibitem{ha2016hypernetworks}
D.~Ha, A.~Dai, and Q.~V. Le,
\newblock ``Hypernetworks,''
\newblock {\em arXiv preprint arXiv:1609.09106}, 2016.

\bibitem{ha2017neural}
D.~Ha and D.~Eck,
\newblock ``A neural representation of sketch drawings,''
\newblock {\em arXiv preprint arXiv:1704.03477}, 2017.

\bibitem{ljspeech17}
K.~Ito,
\newblock ``The lj speech dataset,''
  \url{https://keithito.com/LJ-Speech-Dataset/}, 2017.

\bibitem{gentle17}
R.~M. Ochshorn and M.~Hawkins,
\newblock ``Gentle forced aligner [computer software],'' 2017,
\newblock https://github.com/lowerquality/gentle.

\end{thebibliography}

\end{document}